\documentclass[lettersize,journal]{IEEEtran}
\usepackage{amsmath,amsfonts}
\usepackage{algorithmic}
\usepackage{algorithm}
\usepackage{array}
\usepackage[caption=false,font=normalsize,labelfont=sf,textfont=sf]{subfig}
\usepackage{textcomp}
\usepackage{stfloats}
\usepackage{url}
\usepackage{verbatim}
\usepackage{graphicx}
\usepackage{cite}
\begin{document}
\title{Unified End-to-End V2X Cooperative Autonomous Driving}
\author{Zhiwei Li\textsuperscript{1},
  Bozhen Zhang\textsuperscript{1}\textsuperscript{*},Lei Yang,
  Tianyu Shen\textsuperscript{*},
  Nuo Xu,
  Ruosen Hao,
  Weiting Li,
  Tao Yan,
  Huaping Liu 
\thanks{\textsuperscript{1} Zhiwei Li and Bozhen Zhang contributed equally to this work.}
\thanks{Corresponding authors:Bozhen Zhang and Tianyu Shen.}
\thanks{Zhiwei Li is with the College of Information Science and Technology, Beijing, 100029,China(e-mail:lizw@buct.edu.cn).}
\thanks{Bozhen Zhang is with the School of Information Science and
Engineering, Shenyang University of Technology, Shenyang, 110870,China(e-mail:siriuszhang025@gmail.com).}  
\thanks{Lei Yang is with the School of Vehicle and Mobility, Tsinghua University, Beijing, 100084,China(e-mail:yanglei20@mails.tsinghua.edu.cn).}
\thanks{Tianyu Shen is with College of Information Science and Technology, Beijing University of Chemical Technology, Beijing, 100029,China(e-mail:tianyu.shen@buct.edu.cn).}
\thanks{Nuo Xu is with School of Software, Beihang University, Beijing, 100191,China.}
\thanks{Ruosen Hao is with the College of Information Science and Technology, Beijing University of Chemical Technology, Beijing, 100029,China.}
\thanks{Weiting Li is with the School of Information and
Engineering, China University of Geosciences Beijing, Beijing, 100083,China.}
\thanks{Tao Yan is with the School of Vehicle and Mobility, Tsinghua University, Beijing, 100084,China.}
\thanks{Huaping Liu is with the Department of Computer Science and Technology, Tsinghua University, Beijing, 100084,China.}}
\markboth{In progress}
{Shell \MakeLowercase{\textit{et al.}}: A Sample Article Using IEEEtran.cls for IEEE Journals}
\maketitle
\begin{abstract}
V2X cooperation, through the integration of sensor data from both vehicles and infrastructure, is considered a pivotal approach to advancing autonomous driving technology. Current research primarily focuses on enhancing perception accuracy, often overlooking the systematic improvement of accident prediction accuracy through end-to-end learning, leading to insufficient attention to the safety issues of autonomous driving. To address this challenge, this paper introduces the UniE2EV2X framework, a V2X-integrated end-to-end autonomous driving system that consolidates key driving modules within a unified network. The framework employs a deformable attention-based data fusion strategy, effectively facilitating cooperation between vehicles and infrastructure. The main advantages include: 1) significantly enhancing agents' perception and motion prediction capabilities, thereby improving the accuracy of accident predictions; 2) ensuring high reliability in the data fusion process; 3) superior end-to-end perception compared to modular approaches. Furthermore, We implement the UniE2EV2X framework on the challenging DeepAccident, a simulation dataset designed for V2X cooperative driving.
\end{abstract}
\begin{IEEEkeywords}
V2X Cooperation, End-to-End, Autonomous Driving
\end{IEEEkeywords}
\section{Introduction}
\IEEEPARstart{O}{ver} the past few decades, the transportation\cite{yu2021automated} and automotive sectors\cite{chen2022milestones}\cite{yurtsever2020survey} have seen increasing automation and intelligence, driven by advancements in deep learning\cite{9458968}\cite{hagedorn2023rethinking}\cite{kiran2021deep}, control theory\cite{liu2023systematic}\cite{zhang2018novel}, and technologies like sensors\cite{wang2019multi}\cite{huang2022multi} and network communications\cite{ma2023overview}\cite{balkus2022survey}\cite{alladi2022comprehensive}\cite{hawlader2024leveraging}. Autonomous driving research is burgeoning, typically segmenting the complex intelligence system required into multiple subtasks based on different stages of driving, such as perception\cite{li2023delving}\cite{han2023collaborative}, prediction\cite{mozaffari2020deep}, planning\cite{hang2020integrated}, and control\cite{kuutti2020survey}. These multi-stage approaches necessitate maintenance of inter-module communication, potentially leading to system response delays and information loss\cite{zhao2023autonomous}\cite{natan2022fully}. Conversely, end-to-end autonomous driving methods\cite{chib2023recent}\cite{teng2023motion}\cite{coelho2022review} offer a more intuitive and streamlined approach by directly translating environmental data into vehicle control decisions, thus reducing system complexity and minimizing delays through unified data representation.
However, the perception range of individual vehicle intelligence is limited to its onboard sensors, potentially compromising its perception capabilities under complex road and adverse weather conditions. Vehicle-to-Everything (V2X) cooperation\cite{chen2020vision}\cite{gyawali2020challenges}\cite{garcia2021tutorial} enhances autonomous vehicles by integrating information exchange and collaborative operation between vehicles and road infrastructure. This provides comprehensive, accurate road and traffic signal information, improving safety and efficiency. Moreover, V2X communication enables vehicles to perceive beyond their immediate vicinity, facilitating cooperative driving among vehicles.
Despite the focus on improving metrics like detection accuracy and trajectory prediction precision in current V2X research, these improvements do not necessarily equate to effective planning outcomes due to the introduction of irrelevant information by multi-stage autonomous driving methods. This paper proposes an end-to-end V2X-based autonomous driving framework aimed at collision prediction outcomes, integrating target detection, tracking, trajectory prediction, and collision warning into a unified end-to-end V2X autonomous driving method.
The remainder of the paper is organized as follows: Section 2 reviews related work, Section 3 introduces the end-to-end V2X neural network model, Section 4 presents experiments using public datasets and compares them with other models, validating the effectiveness of the proposed method. Section 5 concludes the paper.
\section{Related Work}
\label{sec:Related Work}
\subsection{End-to-End Autonomous Driving}
End-to-end autonomous driving methods are architecturally simpler than modular approaches, directly producing driving commands from perception data, thus avoiding the generation of redundant intermediate stage information. These methods can be implemented through two approaches: imitation learning and reinforcement learning\cite{tampuu2020survey}. Imitation learning, a form of supervised learning, involves learning strategies and updating models by imitating human driving behavior. Its advantage lies in high training efficiency, but it requires extensive annotated training data and cannot cover all potential traffic and driving scenarios. Reinforcement learning, in contrast, learns directly from the interaction between the model and the environment to maximize cumulative rewards. Its advantage is that it does not require manually collected and annotated data, but the model convergence is slow, and the results are significantly affected by the definition of rewards and other factors.
Early end-to-end autonomous driving methods primarily focused on imitation learning, typically outputting simple driving tasks. Using CNNs to infer steering angles from images captured by three cameras, \cite{bojarski2016end} achieved lane keeping on roads without lane markings. Considering vehicle speed, \cite{yang2018end} introduced temporal information on top of CNNs using LSTM, an approach effective for simple tasks like lane keeping but limited in complex traffic scenarios and driving tasks\cite{codevilla2019exploring}. Several studies have implemented end-to-end autonomous driving through reinforcement learning, handling more complex scenarios compared to imitation learning\cite{agarwal2021learning},\cite{ahmed2021policy},\cite{huang2023multi}. Integrating multimodal data into end-to-end autonomous driving models has resulted in better performance than single-modal approaches\cite{xiao2020multimodal},\cite{ye2023fusionad},\cite{chen2021interpretable}. However, the challenge with end-to-end methods lies in poor model interpretability, making it difficult to diagnose and address issues when problems arise. UniAD unifies multiple shared BEV feature-based Transformer networks, containing modules for tracking, mapping, and trajectory prediction. This enhances the model's interpretability, aids in training and troubleshooting, and employs the final planning outcomes to design the loss function, constructing an end-to-end autonomous driving model\cite{hu2023planning}.
\subsection{Vehicle-to-Everything Cooperation}
Autonomous vehicles based on single-vehicle intelligence perceive the environment centered around the vehicle itself using onboard sensors. However, the real-world traffic scene is complex and variable, with single vehicles, especially in terms of perceiving vulnerable road users\cite{yusuf2024vehicle}. Thanks to the advancement of communication technologies, Cooperative Automated Driving Vehicles (CAVs) have been proposed, enhancing the vehicle's perception capabilities by aggregating perception data from other autonomous vehicles in the traffic environment.
Autonomous driving cooperative perception can be categorized into three types based on the data transmitted via V2X. The first type involves direct transmission of raw point cloud data for cooperative perception, demanding high transmission bandwidth\cite{chen2019cooper}. The second approach processes the raw perception data into unified feature information, such as BEV spatial features, before transmission to save bandwidth. This method balances bandwidth requirements and detection accuracy and is the mainstream V2X transmission method\cite{liu2020who2com},\cite{liu2020when2com},\cite{hu2022where2comm},\cite{wang2020v2vnet},\cite{yin2023v2vformer}. The third type generates prediction results for each autonomous vehicle before transmitting this outcome information via V2X, requiring low bandwidth but demanding high accuracy in individual vehicle prediction results\cite{hurl2020trupercept}.
High-quality datasets in autonomous driving cooperative perception have propelled research in the field, with mainstream datasets including V2X-Sim\cite{li2022v2x}, OPV2V\cite{xu2022opv2v}, and DAIR-V2X\cite{yu2023vehicle}. However, these mainstream vehicle-road cooperation datasets primarily focus on perception accuracy as the evaluation metric, suitable for testing the performance of autonomous driving perception algorithms but not for evaluating end-to-end related algorithms. DeepAccident\cite{wang2024deepaccident} is a large-scale autonomous driving dataset generated using the CARLA simulator, supporting testing for end-to-end motion and accident prediction tasks. In this work, we propose an end-to-end autonomous driving framework based on vehicle-road cooperation and utilize the DeepAccident dataset to test the performance of related algorithms.
\section{Methodology}
\label{sec:Methodology}
This paper introduces a vehicle-road cooperative end-to-end autonomous driving framework comprising two major components: the V2X Cooperative Feature Encoder and the End-to-End Perception, Motion, and Accident Prediction Module.
\subsection{V2X Cooperative Feature Encoding Based on Temporal Bird's Eye View}
\paragraph{Overall Structure}
Our V2X framework includes both the vehicle itself and road infrastructure. During the cooperative phase, each agent first extracts and converts multi-view image features into BEV features. These are then encoded to align the temporal sequence of V2X agent BEV perception information. Finally, by merging the BEV features of the vehicle with those of the roadside infrastructure, we obtain the cooperative perception features. The process of extracting V2X cooperative features based on temporal BEV, as illustrated in Figure 1, consists of two main components: the multi-view image to BEV feature module based on spatial BEV encoding, and the temporal cascading BEV feature fusion module based on temporal BEV encoding. After spatial transformation and temporal fusion, the infrastructure BEV features are aligned and integrated with the vehicle's coordinate system using a deformable attention mechanism to fuse the two aligned BEV features.enhancing the vehicle's perception capabilities to achieve the final V2X cooperative BEV features.
\begin{figure*}
    \begin{center}			         
        \includegraphics[width=0.9\textwidth]{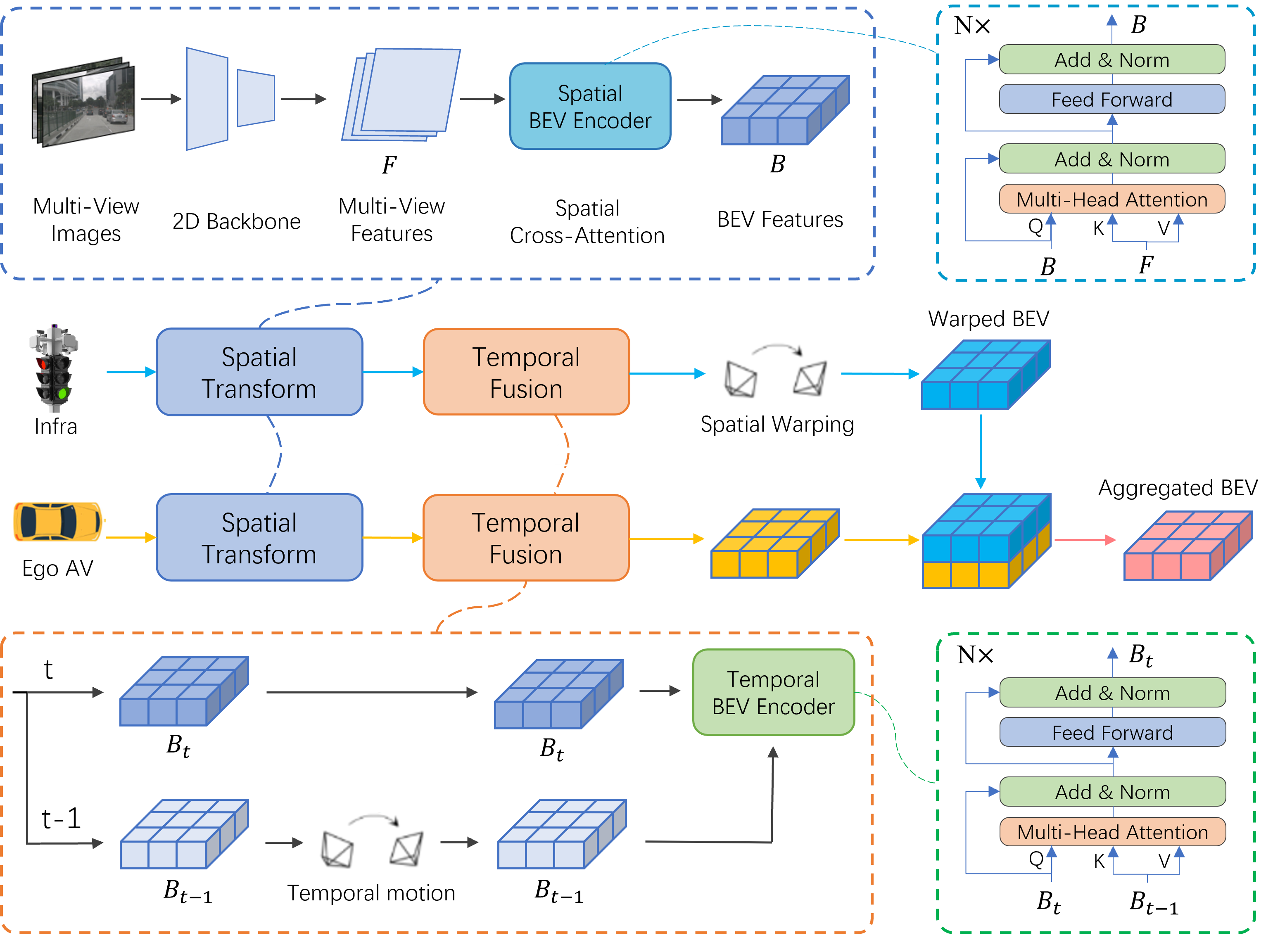} 
        \caption{The overall architecture of V2X collaborative feature extraction process based on time-series BEV}
        \label{fig:V2X}
    \end{center}
\end{figure*}
\paragraph{Multi-View Image to BEV Feature Module Based on Spatial BEV Encoding}
The original perception information obtained from both the vehicle and road infrastructure consists of multi-view perspective images. To eliminate spatial semantic differences and merge multi-source perception data, the multi-view images from both the vehicle and infrastructure are processed through two parallel channels of feature extraction and transformation to yield unified BEV features. Following the approach in \cite{li2022bevformer}, we map multi-view perspective images into the BEV space.
In the module for converting multi-view images to BEV features based on spatial BEV encoding, multi-view images are first processed separately. Two-dimensional convolution is used to extract multi-view feature maps, which are then inputted into the spatial BEV encoder module. The spatial BEV encoder ultimately generates high-level semantic BEV features of the images. This process can be described by Equation (1), where ResNet refers to the ResNet-101 backbone network, $I_{ego}^0$,$I_{ego}^1$,…,$I_{ego}^5$  represent camera images from six viewpoints of the vehicle, and $F_{ego}^0$,$F_{ego}^1$,…,$F_{ego}^5$ represent the feature maps from these six viewpoints. Similarly, $F_{inf}^0$,$F_{inf}^1$,…,$F_{inf}^5$ are the feature maps from six viewpoints of the road infrastructure.
\begin{equation}
F_{ego}^0,F_{ego}^1,…,F_{ego}^5=ResNet(I_{ego}^0,I_{ego}^1,…,I_{ego}^5 )
\end{equation}
Next, the multi-view feature maps are inputted into a spatial BEV encoder based on a deformable spatial cross-attention mechanism to transform two-dimensional image features into BEV spatial features. Initially, a BEV target query $\mathbf{Q} \in \mathbb{R} ^ {H \times W \times C}$, a learnable parameter tensor, is created to gradually learn the BEV information of the multi-view images under the action of the spatial BEV encoder. Q serves as the query for the spatial BEV encoder, with multi-view feature maps $F_{ego}^i$ or $F_{inf}^i$ as the keys and values for the encoder. After six rounds of BEV feature encoding interactions, the parameters of Q are continually updated to yield a complete and accurate BEV feature value B. The specific BEV encoding process can be represented by Equations (2) and (3), where $Q$,$K^i$,$V^i$ respectively denote the BEV target query, image BEV key, and image BEV value. $W^q$,$W^k$,$V^i$ represent the weight matrices for $Q$,$K^i$,$V^i$, and $B$,$F^i$ denote the BEV features and image features, respectively.
\begin{equation}
Q=W^q B,\hspace{2em}      K^i=W^k F^i,\hspace{2em}       V^i=W^v F^i
\end{equation}
\begin{equation}
B=softmax(\frac {Q(K^i)^T} {\sqrt{d}} ) V^i
\end{equation}
However, the query in traditional Transformer architecture encoders conducts attention operations with all keys, which is neither efficient nor necessary given the vast scale and mixed signals of multi-view feature maps serving as keys. Hence, in actual BEV feature encoding, encoders based on a deformable attention mechanism are used to conserve computational resources and enhance efficiency significantly.
\paragraph{Temporal Cascading BEV Feature Fusion Module Based on Temporal BEV Encoding}
The BEV features \(B_t\) obtained in the previous section are considered carriers of sequential information. Each moment's BEV feature \(B_t\) is based on the BEV feature from the previous moment \(B_{(t-1)}\) to capture temporal information. This approach allows for the dynamic acquisition of necessary temporal features, enabling the BEV features to more quickly and effectively respond to changes in the dynamic environment.
In the temporal cascading BEV feature fusion module based on temporal BEV encoding, the BEV feature from the preceding frame \(B_{(t-1)}\) serves as prior information to enhance the current frame's BEV feature \(B_t\). Since \(B_t\) and \(B_{(t-1)}\) are in their respective vehicle coordinate systems, the \(B_{(t-1)}\) feature must first be transformed to the current frame \(B_t\)'s vehicle coordinate system using the vehicle's position transformation matrix. Then, \(B_t\) and \(B_{(t-1)}\), as two frames of BEV features, are inputted, and a temporal BEV encoder based on a deformable cross-attention mechanism is used to transform two-dimensional image features into cooperative perception BEV features.
First, static scene alignment is achieved. Knowing the world coordinates of the vehicle at moments \(t-1\) and \(t\), and using the continuous frame vehicle motion transformation matrix, \(B_{(t-1)}\) features are aligned to \(B_t\). This alignment operation ensures that \(B_{(t-1)}\) and \(B_t\) in the same search position grids correspond to the same location in the real world, with the aligned BEV features denoted as \(B_{(t-1)}'\).
Subsequently, dynamic target alignment is executed. The BEV feature \(B_t\) at time \(t\) serves as the target query \(Q \in \mathbb{R}^{H \times W \times C}\), progressively learning the BEV features of time \(t-1\) under the action of the temporal BEV encoder. \(Q\) is used as the query for the temporal BEV encoder, with the previous moment's BEV features serving as keys and values. Through BEV feature encoding interactions, \(Q\)'s parameters are continuously updated, ultimately yielding a complete and accurate cooperative perception BEV feature value \(B_t\). The specific BEV encoding process is represented by Equations (4) and (5), where \(Q\), \(K^i\), and \(V^i\) respectively represent the target query for BEV features at time \(t\), the key for image BEV features at time \(t-1\), and the value for image BEV features at time \(t-1\). \(W^q\), \(W^k\), and \(V^i\) are the weight matrices for \(Q\), \(K^i\), and \(V^i\), with \(B_t\) and \(B_{(t-1)}\) representing BEV features at time \(t\) and image BEV features at time \(t-1\), respectively.
\begin{equation}
Q = W^q B_t, \quad K^i = W^k B_{(t-1)}, \quad V^i = W^v B_{(t-1)}
\end{equation}
\begin{equation}
B_t = \text{softmax}\left(\frac{Q(K^i)^T}{\sqrt{d}}\right) V^i
\end{equation}
At time \(t-1\), assuming a target is present at some point in \(B_{(t-1)}\), it is likely that the target will appear near the corresponding point in \(B_t\) at time \(t\). By employing the deformable cross-attention mechanism focusing on this point and sampling features around it, high-precision temporal feature extraction with low overhead can be achieved in dynamic and complex environments.
\subsection{End-to-End Autonomous Driving}
We propose a unified end-to-end V2X cooperative autonomous driving model named UniE2EV2X, oriented towards accident prediction. The primary tasks of this model include object detection and tracking, motion prediction, and post-processing for accident prediction, as illustrated in Figure 2.
\begin{figure*}
    \begin{center}			         
        \includegraphics[width=0.9\textwidth]{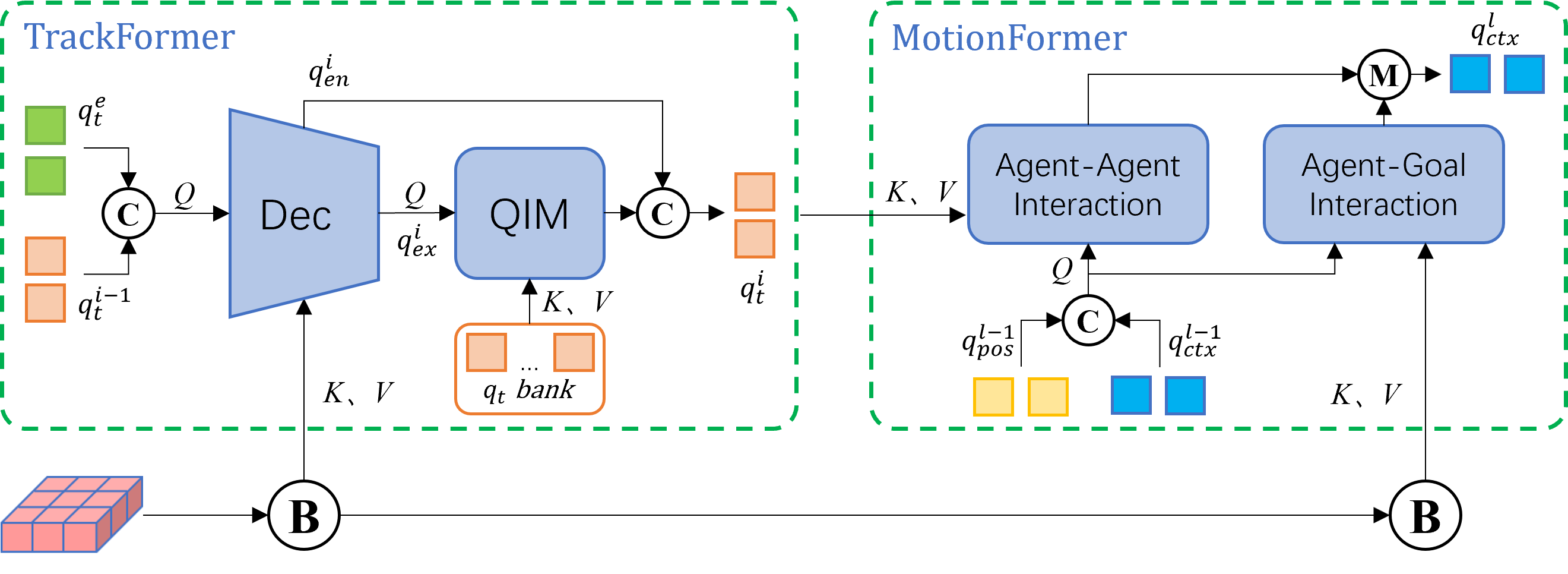}
        \caption{The tasks of UniE2EV2X}
        \label{fig:V2X}
    \end{center}
\end{figure*}
\paragraph{Detection and Track}
The perception module is the initial component of the end-to-end autonomous driving framework presented in this paper. It consists of detection and tracking sub-modules, taking collaborative BEV features as input and producing tracked proxy features for use in the downstream motion prediction module. The detection sub-module is responsible for predicting target information under collaborative BEV features in each time frame, including target locations and dimensions. The tracking sub-module associates the same targets across frames by assigning consistent IDs. In this study, detection and tracking tasks are integrated into a unified multi-object tracking module which first conducts detection queries to identify newly appeared targets, then interacts current frame tracking queries with detection queries from preceding frames to aggregate temporal information, and updates the tracking queries for target tracking in subsequent frames. This multi-object tracking query contains features representing target information over consecutive frames. Additionally, an ego-vehicle query module is introduced to aggregate the trajectory of the self-driving car, which is later used to predict the vehicle’s future trajectory. The multi-object tracking module consists of N Transformer layers, and the output features, $Q_A$, contain rich proxy target information that will be further utilized in the motion prediction module.
\paragraph{Motion Prediction}
The motion prediction module takes the multi-object tracking queries, $Q_A$, and collaborative BEV features from the perception module as inputs. Using a scene-centric approach, it outputs motion queries, $Q_X$, to predict the future trajectories of each proxy and the ego-vehicle over T frames with K possible paths. This method allows simultaneous prediction of multiple proxies' trajectories and fully considers interactions between proxies and between proxies and target locations. The motion queries between proxies, $Q_a$, are derived from multi-head cross-attention mechanisms between motion and tracking queries, while the motion queries related to target locations, $Q_g$, are generated through a variable attention mechanism using motion queries, target positions, and collaborative BEV features. $Q_a$ and $Q_g$ are combined and passed through a multilayer perceptron (MLP) to produce the query context, $Q_ctx$. The motion query positions, $Q_pos$, incorporate four types of positional knowledge: scene-level anchors, proxy-level anchors, the proxies' current positions, and predicted target points. $Q_ctx$ and $Q_pos$ are merged to form the motion query, $Q_X$, which directly predicts each proxy’s motion trajectory.
\paragraph{Accident Prediction}
After inputting collaborative BEV features into the end-to-end autonomous driving framework, the movement predictions for all agents and the ego-vehicle are obtained. These predictions are post-processed frame-by-frame to check for potential accidents. For each timestamp, the predicted motion trajectories of each proxy can be approximated as polygons, and the nearest other targets are identified. By checking if the minimum distance between objects is below a safety threshold, it can be determined whether an accident has occurred, providing labels for the colliding objects’ IDs, positions, and the timestamp of the collision. To assess the accuracy of accident predictions compared to real accident data, the same post-processing steps are applied to actual accident movements to ascertain future accidents' occurrences. The basis for collision includes cases where both predictions and ground truth indicate an accident and the distance between colliding objects is below the threshold.
\section{Experiments}
\label{sec:Experiments}
\section{Conclusion}
\label{sec:Conclusion}
\bibliographystyle{unsrt}  
\bibliography{main}  
\newpage
\end{document}